\documentclass[a4paper]{article}

\usepackage{epsfig}
\usepackage{graphicx}
\usepackage{subcaption}
\usepackage{calc}
\usepackage{amssymb}
\usepackage{amstext}
\usepackage{amsmath}
\usepackage{amsthm}
\usepackage{multicol}
\usepackage{pslatex}
\usepackage{apalike}
\usepackage[bottom]{footmisc}
\usepackage{algorithm}
\usepackage{algorithmic}
\usepackage{multirow}
\usepackage{array}
\usepackage{hyperref}
\usepackage{SCITEPRESS}     

\begin{document}

\title{BBBD: Bounding Box Based Detector for Occlusion Detection and Order Recovery}

\author{\authorname{Kaziwa Saleh\sup{1}\orcidAuthor{0000-0003-3902-1063}, Zolt\'{a}n V\'{a}mossy\sup{2}\orcidAuthor{0000-0002-6040-9954}}
\affiliation{\sup{1}Doctoral School of Applied Informatics
and Applied Mathematics, \'{O}buda University, Budapest, Hungary}
\affiliation{\sup{2}John von Neumann Faculty of
Informatics,\'{O}buda University, Budapest, Hungary}
\email{kaziwa.saleh@stud.uni-obuda.hu, vamossy.zoltan@nik.uni-obuda.hu}
}


\abstract{Occlusion handling is one of the challenges of object detection and segmentation, and scene understanding. Because objects appear differently when they are occluded in varying degree, angle, and locations. Therefore, determining the existence of occlusion between objects and their order in a scene is a fundamental requirement for semantic understanding. Existing works mostly use deep learning based models to retrieve the order of the instances in an image or for occlusion detection. This requires labelled occluded data and it is time-consuming. In this paper, we propose a simpler and faster method that can perform both operations without any training and only requires the modal segmentation masks. For occlusion detection, instead of scanning the two objects entirely, we only focus on the intersected area between their bounding boxes. Similarly, we use the segmentation mask inside the same area to recover the depth-ordering. When tested on COCOA dataset, our method achieves +8\% and +5\% more accuracy than the baselines in order recovery and occlusion detection respectively.}

\onecolumn \maketitle \normalsize \setcounter{footnote}{0} \vfill

\section{\uppercase{Introduction}}
\label{sec:introduction}

Real-world scenes are complex and cluttered, as humans we observe and fathom them effortlessly even when objects are not fully visible. We can easily deduce that an object is partially hidden by other objects. For machines, this is a challenging task particularly if the object(s) are occluded by more than one object. Nevertheless, for a machine to comprehend its surrounding, it has to be capable of inferring the order of objects in the scene, i.e. to determine if the object is occluded and by which object(s). 

\begin{figure*}[t]
		\begin{subfigure}[t]{0.3\textwidth}
			\includegraphics[width=\textwidth]{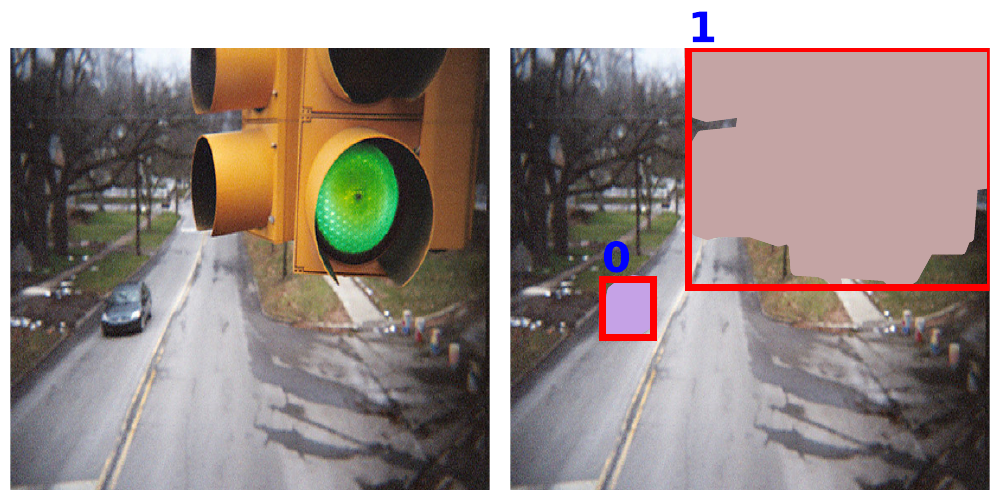}
			\caption{}
		\end{subfigure}
		\hfill
		\begin{subfigure}[t]{0.3\textwidth}
			\includegraphics[width=\textwidth]{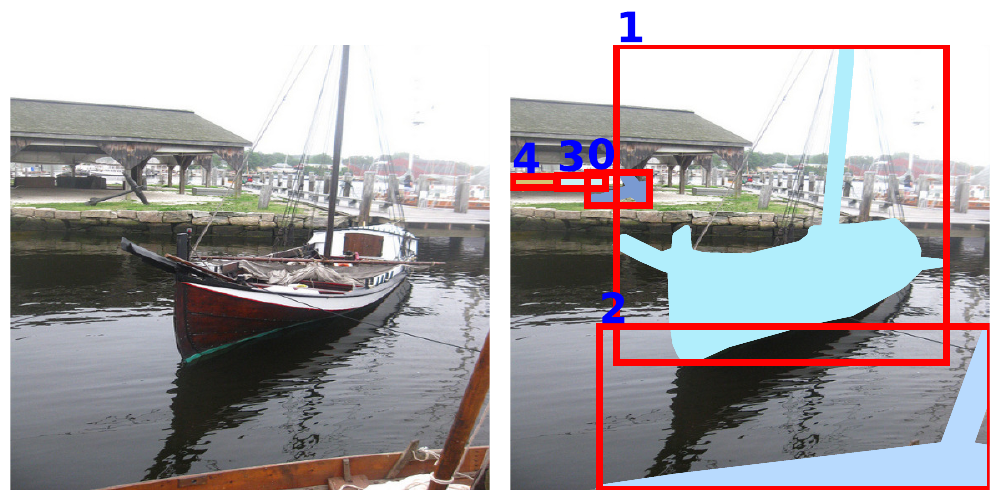}
			\caption{}
		\end{subfigure}
		\hfill
		\begin{subfigure}[t]{0.3\textwidth}
			\includegraphics[width=\textwidth]{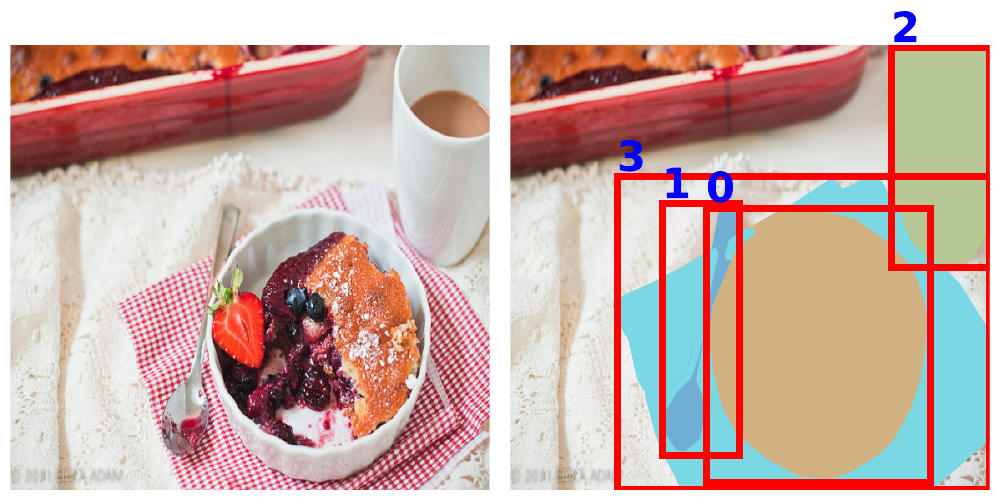}
			\caption{}
		\end{subfigure}
		
		\begin{subfigure}[t]{0.3\textwidth}
			\includegraphics[width=\textwidth]{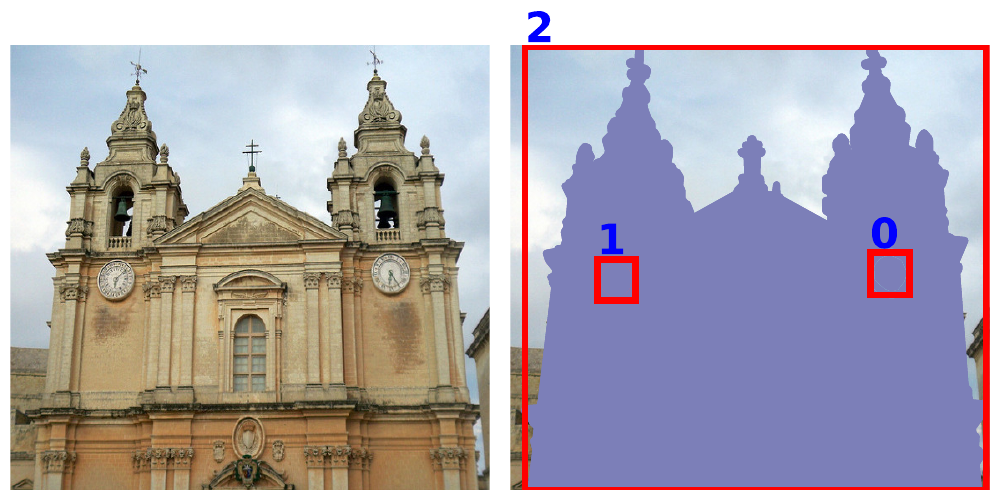}
			\caption{}
		\end{subfigure}
		\hfill
		\begin{subfigure}[t]{0.3\textwidth}
			\includegraphics[width=\textwidth]{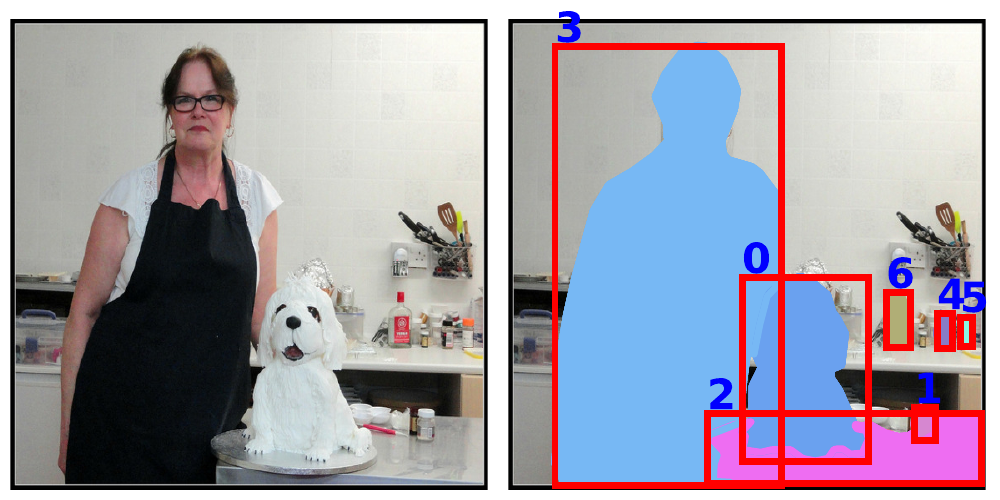}
			\caption{}
		\end{subfigure}
		\hfill
		\begin{subfigure}[t]{0.3\textwidth}
			\includegraphics[width=\textwidth]{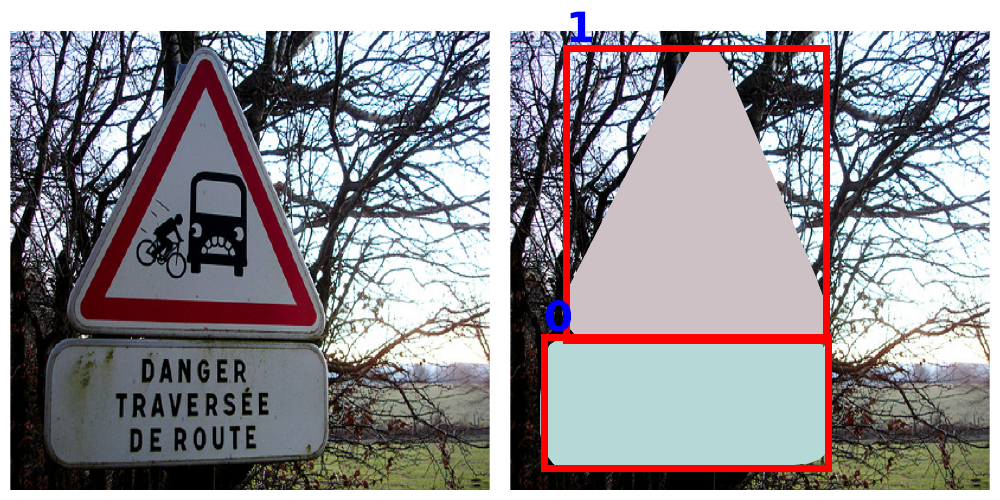}
			\caption{}
		\end{subfigure}
		\caption{Cases considered in BBBD: (a): IA is zero, (b): IA of bounding boxes 1 and 2 is non-zero, however it does not contain mask from 2, (c) masks from bounding boxes 1 and 2 do not collide inside the IA, (d) bounding boxes 0 and 1 are inside 2, (e) object in bounding box 0 is the occluder since it has bigger area inside the IA with 2, (f) IA is zero.}
		\label{fig:cases}
	\end{figure*}
	
Working with occlusion is difficult because an object can be hidden by other object(s) in varying ratio, location, and angle. Yet, handling it plays a key role in the machine perception. Many works in the recent literature address occlusion in various applications \cite{saleh2021review,toth2020tree}. The focus is either on detecting and segmenting the occluded object \cite{ke2021deep} \cite{zheng2021visiting}, completing the invisible region \cite{Wang_2021_ICCV} \cite{ling2020variational}, or depth-ordering \cite{zhan2020self}\cite{ehsani2018segan}. 

However, almost all the works in the literature rely on deep learning based network to retrieve the amodal mask (the mask for occluded region) of the object and utilize it for occlusion presence detection and depth ordering. Although this produces good results, it is time-consuming and it requires a labelled occluded dataset which may not be available \cite{finta2019state}.  

In this work, we propose a simpler approach. The method only requires the modal masks and their bounding boxes which can easily be obtained. In contrast to scanning the entire mask of objects to determine if they occlude each other or not, we only focus on the intersected area (IA) of the bounding boxes. We utilize the portion of the mask that falls into this area. Since we concentrate on a smaller region, our method is faster and can produce instant results.

Our method called Bounding Box Based Detector (BBBD) takes the bounding boxes and modal masks of the instances detected in the image as input, and outputs a matrix that contains the order of the objects. By calculating the IA of the bounding boxes and checking the mask in that area, we infer the existence of occlusion. 
The method is tested on COCOA \cite{zhu2017semantic} dataset, and the achieved results are higher than the baselines. This makes BBBD suitable for integration with other deep learning based models, particularly for occlusion detection.

In summary, this work contributes as follows: 1) We propose BBBD, a simple and fast method that requires only the bounding boxes and the modal masks of the objects in an image, to recover the depth order of the objects. 2) By calculating the IA between bounding boxes and only checking the modal mask in that area, we can also determine the presence of occlusion. 3) BBBD can achieve higher accuracy in order recovery and occlusion detection when compared against the baselines. 4) The method does not require any training or amodal segmentation mask, which makes it easier to use.

\section{\uppercase{Related Work}}

\textbf{Occlusion Detection:} Li and Malik in \cite{ke2016amodal} propose an Iterative Bounding Box Expansion method which predicts the amodal mask by iteratively expanding the amodal segmentation in the pixels whose intensities in the heatmap is above a threshold. Then by using the predicted modal and amodal segmentation masks, they compute the area ratio which shows how much an object is occluded.

Qi et al. in \cite{qi2019amodal} propose Multi-Level Coding (MLC) network that combines the global and local features to produce the full segmentation mask, and predicts the occlusion existence through an occlusion classification branch.
\\
\noindent \textbf{Order Recovery:} Ehsani et al. \cite{ehsani2018segan} recover the full mask of an object and use it to infer the depth order from the depth relation between the occluder and occludee. In \cite{yang2011layered}, authors propose a layered object detection and segmentation method in order to predict the order of the detected objects. \cite{tighe2014scene} predict the semantic label for each pixel and order the objects based on inferred occlusion relationship. \cite{zhan2020self} use a self-supervised model to partially complete the mask of objects in an image. For any two neighbouring instances, the object which requires more completion is considered to be an occludee. \cite{zhu2017semantic} manually annotate a dataset with occlusion ordering and segmentation mask for the invisible regions. Then they rely on the amodal mask to predict the depth ordering.\\

In contrast to the above-mentioned methods, BBBD does not require any training or amodal mask. It merely relies on the modal mask and the bounding boxes to retrieve the order of the instances and predict the existence of occlusion.

\section{\uppercase{Methodology}}

In this section, we describe how BBBD works. Once the bounding boxes and the modal segmentation masks are detected using any off-shelf detector (e.g. Detectron2 \cite{wu2019detectron2}), for any two bounding boxes we find the IA. We can determine that there is no occlusion if any of these cases were true:
\begin{itemize}
    \item IA is zero (see cases 'a' and 'f' in figure \ref{fig:cases}).
    \item IA is non-zero, however, one or both objects’ mask is zero in that area. It means that although the bounding boxes intersect, the IA does not actually contain any parts of the object(s) (case 'b' in figure \ref{fig:cases}).
    \item IA is non-zero, but the objects’ mask do not collide (case 'c' in figure \ref{fig:cases}.
\end{itemize}

\noindent If none of the above cases is true, and IA is non-zero then we conclude that there is occlusion. In the case of occlusion presence, we perform the following checks to obtain the order matrix:
\begin{itemize}
    \item If one of the bounding boxes is fully contained in another one, and both masks in that area are non-zero, then the object with a bigger bounding box is considered the occludee and the smaller one as the occluder (case 'd' in figure \ref{fig:cases}).
    \item We count the number of non-zero pixels in each mask inside the IA, the object with a larger mask is the occludee (case 'e' in figure \ref{fig:cases}).
\end{itemize}
Algorithm \ref{alg:occlusion} describes BBBD in detail.

\begin{algorithm}[h]
\begin{algorithmic} 
\REQUIRE Bounding box and masks of instances in the image: $B_{0:N}$, and $M_{0:N}$
\FOR{$i \leftarrow 0 : N$} 
\FOR{$j \leftarrow i : N$}
\STATE $OrderMatrix \leftarrow $a zero matrix with size $i \times j$
\STATE $IA = B_i \cap B_j$
\IF{$IA \neq 0$}
\STATE $MB_1$ and $MB_2 \leftarrow$ values inside IA region in $M_i$ and $M_j$
\STATE $MC_1$, $MC_2 \leftarrow $ number of pixels in $MB_1$ and $MB_2$
\IF{$MC_1$ or $MC_2$ = 0}
\STATE skip to the next iteration
\ENDIF
\IF{$MB_1$ and $MB_2$ do not collide}
\STATE skip to the next iteration
\ENDIF
\IF{$B_i$ is inside $B_j$}
\STATE $OrderMatrix[i][j] \leftarrow 1$
\STATE $OrderMatrix[j][i] \leftarrow -1$
\ELSIF{$B_j$ is inside $B_i$}
\STATE $OrderMatrix[i][j] \leftarrow -1$
\STATE $OrderMatrix[j][i] \leftarrow 1$
\ENDIF
\IF{$MC_1 > MC_2$}
\STATE $OrderMatrix[i][j] \leftarrow 1$
\STATE $OrderMatrix[j][i] \leftarrow -1$
\ELSIF{$MC_1 < MC_2$}
\STATE $OrderMatrix[i][j] \leftarrow -1$
\STATE $OrderMatrix[j][i] \leftarrow 1$
\ENDIF
\ENDIF
\ENDFOR
\ENDFOR
\RETURN $OrderMatrix$
\end{algorithmic}
\caption{Pseudocode for BBBD for occlusion existence detection and order recovery.}
\label{alg:occlusion}
\end{algorithm}

\section{\uppercase{RESULTS}}
We compared our method to two baselines, Area (the bigger object is the occluder, and the smaller is the occludee), and Y-axis (The instance with larger Y value is considered to be the occluder) \cite{zhu2017semantic}. In the following sections we present the results for occlusion presence detection and order recovery separately.

	\begin{figure*}[!ht]  
	   \centering
		\begin{subfigure}[t]{1.0\textwidth}
			\includegraphics[width=\textwidth]{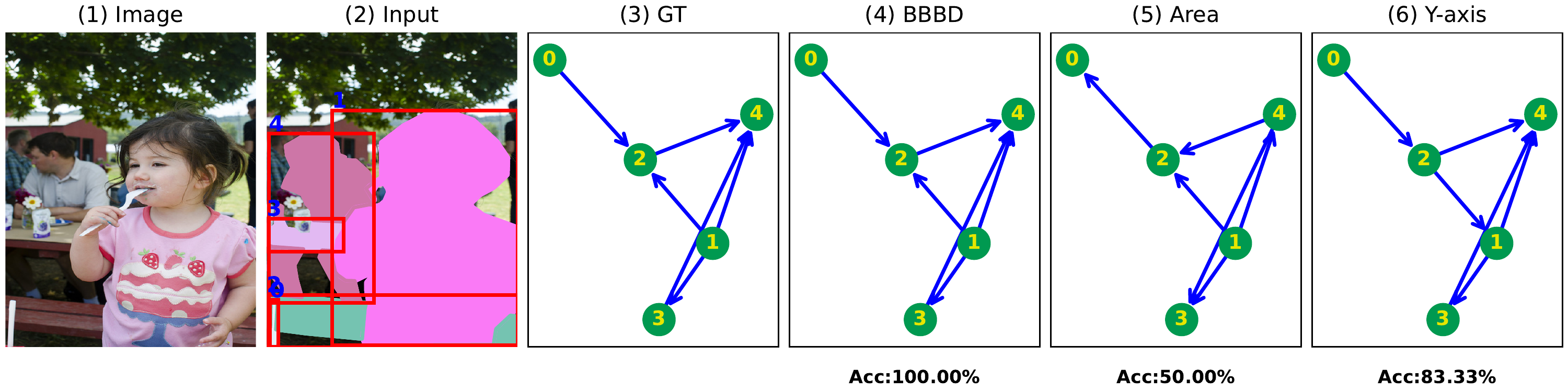}
		\end{subfigure}
	
		\begin{subfigure}[t]{1.0\textwidth}
			\includegraphics[width=\textwidth]{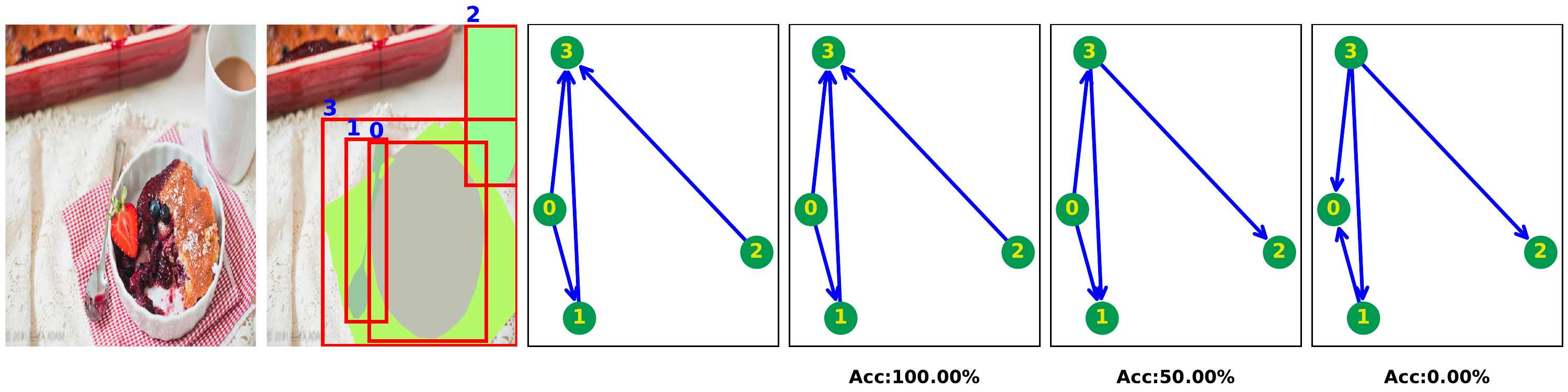}
		\end{subfigure}
	
		\begin{subfigure}[t]{1.0\textwidth}
			\includegraphics[width=\textwidth]{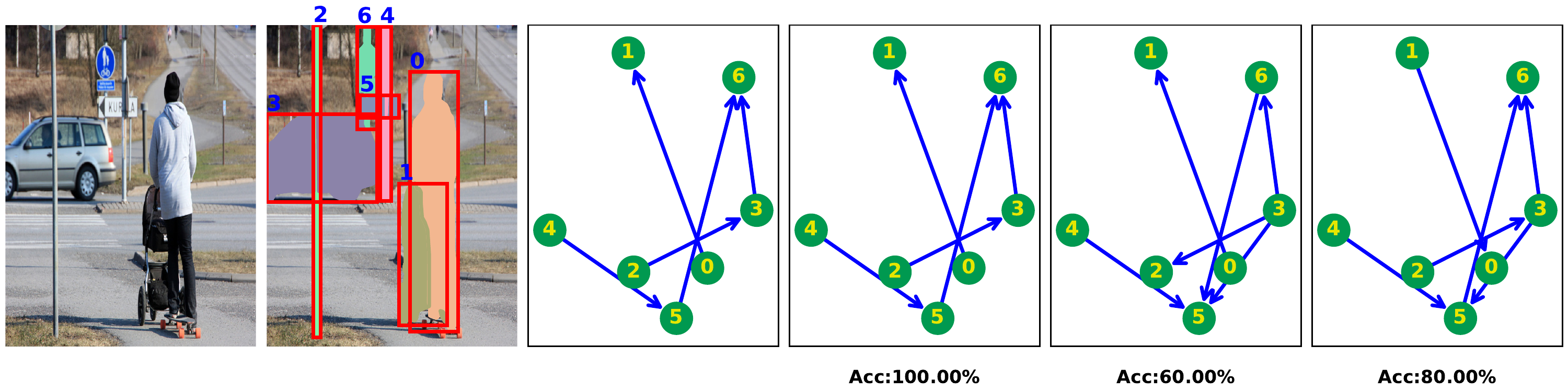}
		\end{subfigure}
	
		\begin{subfigure}[t]{1.0\textwidth}
			\includegraphics[width=\textwidth]{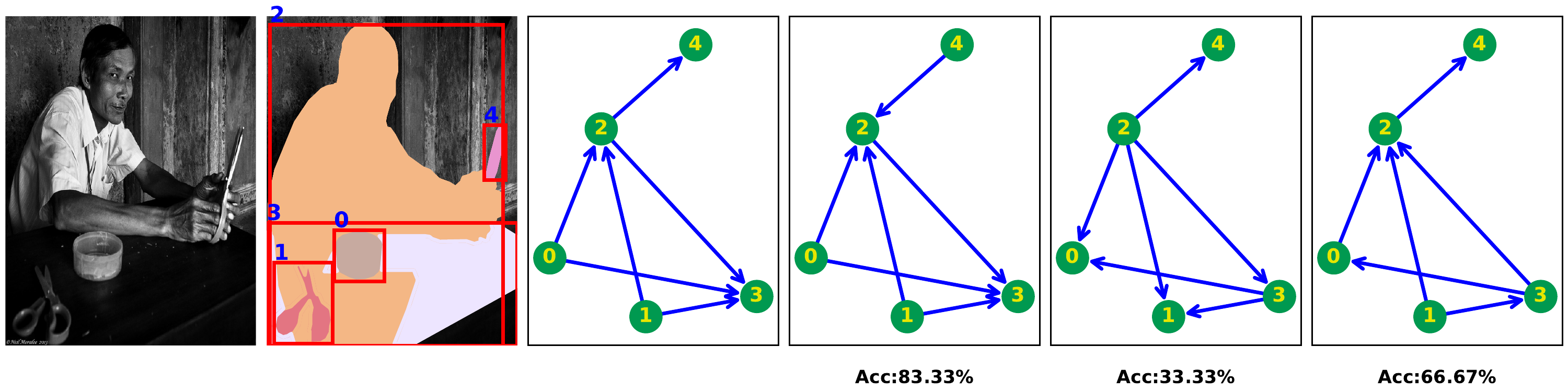}
		\end{subfigure}

		\caption{Results for the order recovery. The nodes depict the detected objects. The directed edges show the occlusion relationship between the objects. The occludee and the occluder are described by the source and the target nodes, respectively.}
		\label{fig:goodResults}
	\end{figure*}

	\begin{figure*}[!ht]  
	    \centering
		\begin{subfigure}[t]{1.0\textwidth}
			\includegraphics[width=\textwidth]{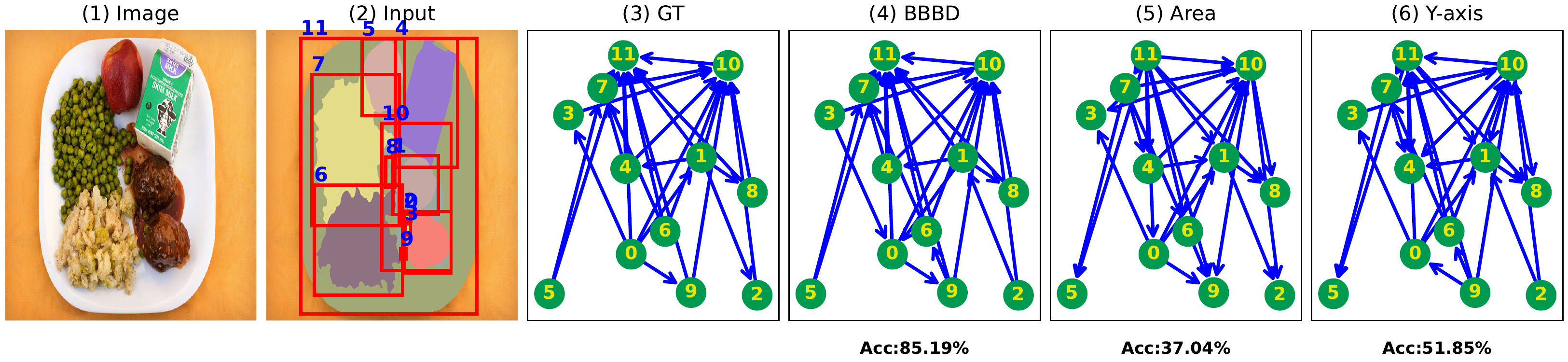}
		\end{subfigure}
	
		\begin{subfigure}[t]{1.0\textwidth}
			\includegraphics[width=\textwidth]{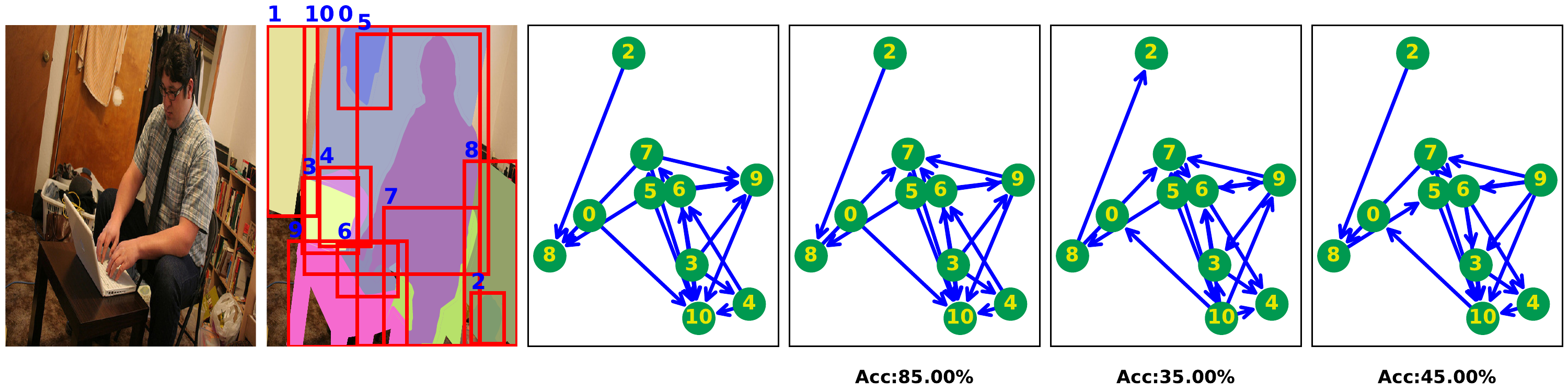}
		\end{subfigure}
	
		\begin{subfigure}[t]{1.0\textwidth}
			\includegraphics[width=\textwidth]{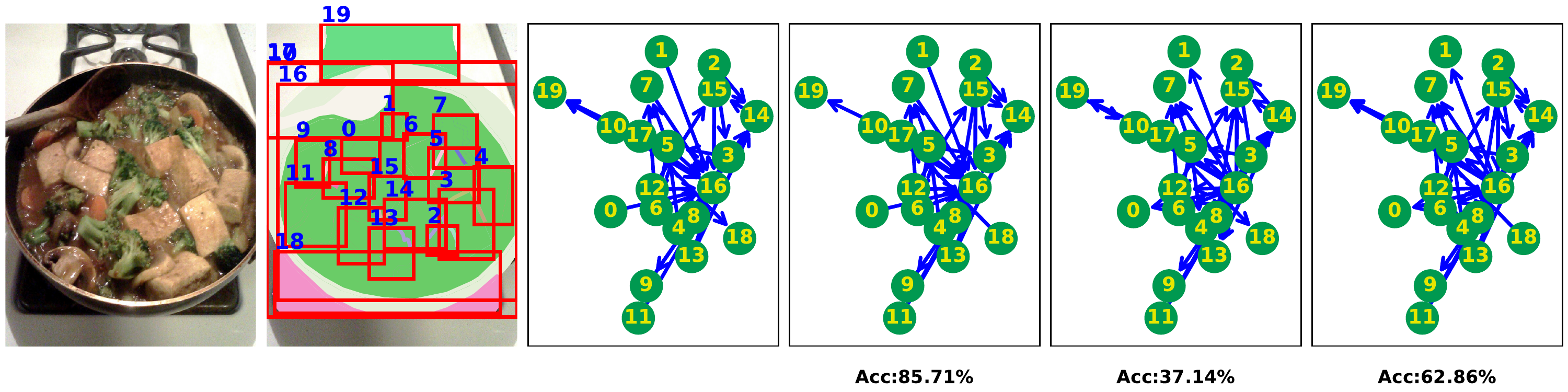}
		\end{subfigure}
		\caption{Results for crowded scenes.}
		\label{fig:crowdedScene}
	\end{figure*}
	\begin{figure*}[!ht] 
	    \centering
		\begin{subfigure}[t]{1.0\textwidth}
			\includegraphics[width=\textwidth]{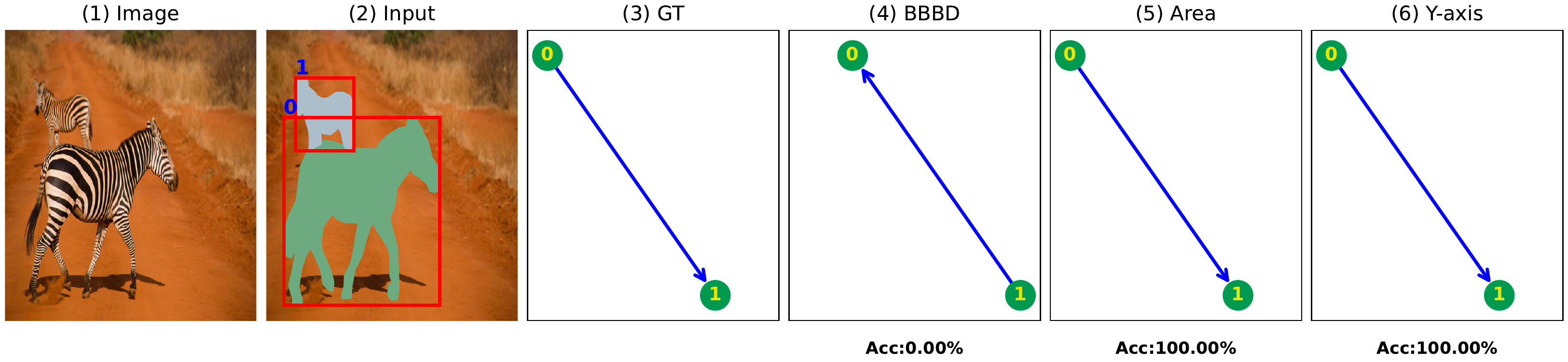}
		\end{subfigure}

		\begin{subfigure}[t]{1.0\textwidth}
			\includegraphics[width=\textwidth]{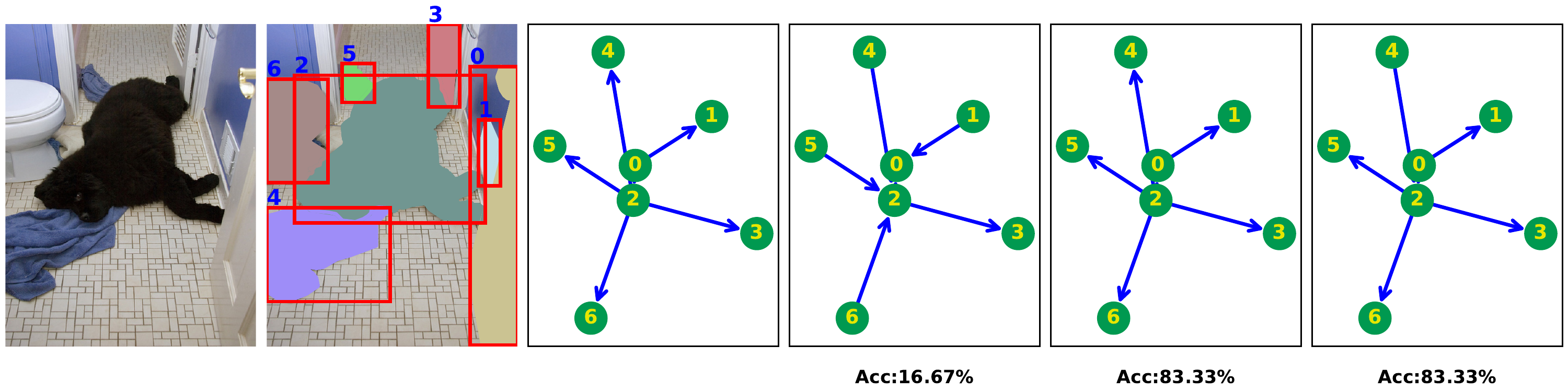}
		\end{subfigure}

		\begin{subfigure}[t]{1.0\textwidth}
			\includegraphics[width=\textwidth]{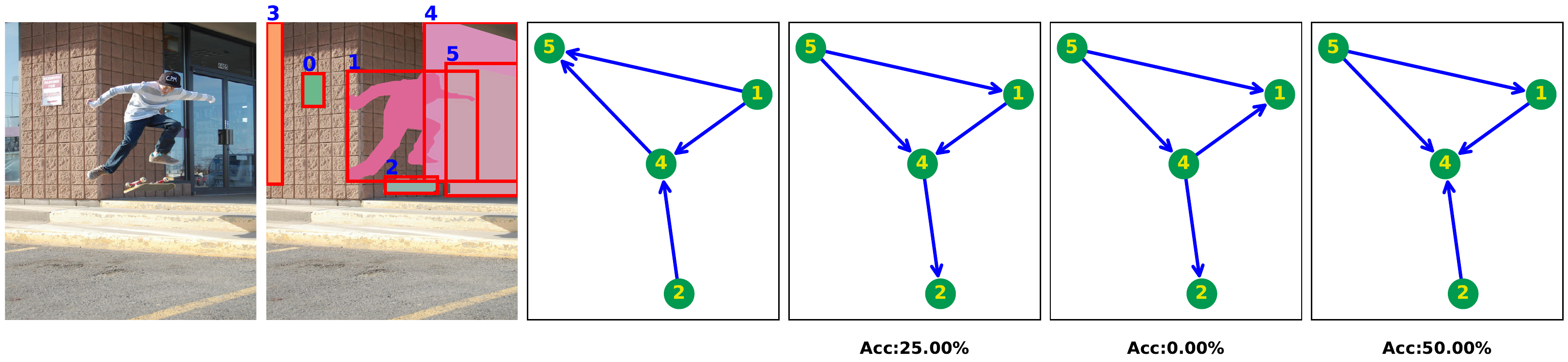}
		\end{subfigure}

		\caption{Results where our algorithm fails.}
		\label{fig:badResults}
	\end{figure*}

\subsection{Occlusion Detection}
To test our method in occlusion existence detection, we relied on the order matrix that we obtained after the order recovery. The size of the matrix is based on the number of detected objects in the image. The value of a cell is '-1' if the instance in the corresponding row is occluded by the instance in the column. If the instance is the occluder, the value in the cell is '1'. Otherwise, it is '0'. 

For each instance, if the corresponding row contains no '-1' values then we determine that that is no occlusion for that instance. The values '0' or '1' show that the object is either isolated or it is the occluder, respectively.

The same approach is used to evaluate the results from the order matrix obtained from the two baselines.

\begin{table}[h]
\caption{Accuracy and precision results for the occlusion detection.}\label{tab:accOcc} \centering
\begin{tabular}{|c|c|c|c|}
  \hline
  &\textbf{BBBD}&\textbf{Y-axis}&\textbf{Area}\\
  \hline
  \textbf{Accuracy} & \textbf{73.05\%} & 68.56\% & 65.16\% \\
  \hline
  \textbf{Precision} & \textbf{78.49\%} & 70.09\% & 66.50\% \\
  \hline
  \textbf{Recall} & 69.99\% & \textbf{74.33}\% & 73.31\% \\
  \hline
\end{tabular}
\end{table}

The method is tested on COCOA validation set, which contains the occlusion ratio for 1323 images. The dataset in total includes 9508 objects, with 4300 negative samples and 5208 positive samples. Table \ref{tab:accOcc} illustrate that BBBD achieves better accuracy in detecting the occlusion presence. 

\begin{table}[h]
\caption{Confusion matrix for the occlusion detection.The percentage is calculated from the total of 9508 objects}\label{tab:confMatrix} \centering
\begin{tabular}{|c|c|c|c|c|}
  \hline
  &\textbf{TP}&\textbf{FP}&\textbf{TN}&\textbf{FN}\\
  \hline
  \textbf{BBBD} & 38.3\% & 10.5\% & 34.7\% & 16.4\% \\
  \hline
  \textbf{Y-axis} & 40.7\% & 17.4\% & 27.9\% & 14.1\% \\
  \hline
  \textbf{Area} & 40.2\% & 20.2\% & 25.0\% & 14.6\% \\
  \hline
\end{tabular}
\end{table}

As the results of occlusion detection depend on the order matrix, any false prediction in the occlusion order leads to false outcomes of occlusion detection. This explains the results from table \ref{tab:confMatrix}. The table shows that BBBD has less true positive and more false negative predictions by only 2.4\% and 1.8\% compared to Y-axis and Area algorithms, respectively. However, it has less false positive and more true negative predictions by 6.9\% and 9.7\% compared to Y-axis and Area algorithms, respectively. 

\subsection{Order Recovery}
The results in the obtained order matrix are evaluated against the ground truth order matrix from COCOA validation dataset. Table \ref{tab:AccOrd} illustrates that our method surpasses the baselines. Figure \ref{fig:goodResults} visualizes the order matrix from BBBD and the baselines (the plots are created by heavily building on the GitHub repository of \cite{zhan2020self}'s paper). Similarly, figure \ref{fig:crowdedScene} shows that our method achieves acceptable results even in cluttered scenes.
 
\begin{table}[h]
\caption{Accuracy results for the order recovery.}\label{tab:AccOrd} \centering
\begin{tabular}{|c|c|c|}
  \hline
  \textbf{BBBD}&\textbf{Y-axis}&\textbf{Area}\\
  \hline
  69.53\% & 65.43\% & 61.36\% \\
  \hline
\end{tabular}
\end{table}

\subsection{Limitations}
Although the previous results shows that BBBD can give higher accuracy compared to the baselines, in some cases the method does not perform as well as expected. Occlusion ordering by comparing the area inside the IA does not always produce correct output. Since the two tasks of depth ordering and occlusion detection are related, the result of one affects the other. And because there is no other method to evaluate the result of occlusion detection, we had to rely on order matrix to assess the accuracy. Otherwise, we believe the results would have been higher for occlusion detection. Figure \ref{fig:badResults} shows some examples where our method fails to give correct predictions.

\section{\uppercase{Conclusions}}
\label{sec:conclusion}
Due to the significance of occlusion handling in machine vision, determining the occlusion ordering and predicting its presence is essential. These two tasks are related and depend on each other. In this paper, we presented a simple and fast method that can deduce the existence of occlusion between objects in a scene and retrieve their ordering. The method depends only on the bounding box and the modal masks of the objects. From the quantitative results we conclude that BBBD has higher accuracy than its baselines. The method is simple to implement as it does not require any training. However, it can easily be integrated and used with other deep learning methods. Therefore, in the future we plan to use this approach with a trained model to retrieve the occlusion order to obtain optimal results.

\section*{\uppercase{Acknowledgements}}
We acknowledge the support of the Doctoral School of Applied Informatics and Applied Mathematics at the \'{O}buda University, both research groups: the GPGPU Programming and the Applied Machine Learning at the \'{O}buda University, and the '2020-1.1.2-PIACI-KFI-2020-00003' project.

We would also like to thank NVIDIA Corporation for providing graphics hardware through the CUDA Teaching Center program. On the behalf of "Occlusion Handling in Object Detection", we thank the usage of ELKH Cloud (https://science-cloud.hu/) that significantly helped us achieving the results published in this paper.

\bibliographystyle{apalike}

\end{document}